\documentclass{article}

\usepackage{arxiv}

\usepackage[utf8]{inputenc} 
\usepackage[T1]{fontenc}    
\usepackage{hyperref}       
\usepackage{url}            
\usepackage{booktabs}       
\usepackage{amsfonts}       
\usepackage{nicefrac}       
\usepackage{microtype}      
\usepackage{lipsum}		
\usepackage{graphicx}
\usepackage{doi}
\usepackage[numbers]{natbib}
\usepackage{amsmath}
\usepackage{multirow}

\title{Soft insoles for estimating 3D ground reaction forces using 3D printed foam-like sensors}

\author{ \href{}{\hspace{1mm} Nick~Willemstein}\\
	Department of Biomechanical Engineering\\
	University of Twente\\
	Enschede, The Netherlands \\
	\texttt{n.willemstein@utwente.nl} \\
	\And
        \href{}{\hspace{1mm} Saivimal~Sridar}\\
	Department of Biomechanical Engineering\\
	University of Twente\\
	Enschede, The Netherlands \\
	\texttt{s.sridar@utwente.nl} \\
	\And
	{\hspace{1mm}Herman~van der Kooij} \\
	Department of Biomechanical Engineering\\
	University of Twente\\
	Enschede, The Netherlands \\
	\texttt{h.vanderkooij@utwente.nl} \\
	\And
	{\hspace{1mm}Ali~Sadeghi} \\
	Department of Biomechanical Engineering\\
	University of Twente\\
	Enschede, The Netherlands \\
	\texttt{a.sadeghi@utwente.nl} \\}

\begin{document}
\maketitle

\begin{abstract}
	Sensorized insoles provide a tool for gait studies and health monitoring during daily life. For users to accept such insoles they need to be comfortable and lightweight. Previous work has already demonstrated that estimation of ground reaction forces (GRFs) is possible with insoles. However, these are often assemblies of commercial components restricting design freedom and customization. Within this work, we investigate using four 3D-printed soft foam-like sensors to sensorize an insole. These sensors were combined with system identification of Hammerstein-Wiener models to estimate the 3D GRFs, which were compared to values from an instrumented treadmill as the golden standard. It was observed that the four sensors behaved in line with the expected change in pressure distribution during the gait cycle. In addition, the identified (personalized) Hammerstein-Wiener models showed the best estimation performance (on average RMS error 9.3\%, $R^2$=0.85 and mean absolute error (MAE) 7\%) of the vertical, mediolateral, and anteroposterior GRFs. Thereby showing that these sensors can estimate the resulting 3D force reasonably well. These results for nine participants were comparable to or outperformed other works that used commercial FSRs with machine learning. The identified models did decrease in estimation performance over time but stayed on average 11.35\% RMS and 8.6\% MAE after a week with the Hammerstein-Wiener model seeming consistent between days two and seven. These results show promise for using 3D-printed soft piezoresistive foam-like sensors with system identification to be a viable approach for applications that require softness, lightweight, and customization such as wearable (force) sensors.
\end{abstract}

\section{Introduction}
With the current boom in the use of advanced health monitoring technologies for both medical and personal use, wearables have become a large part of day-to-day lives \cite{boulos2011smartphones}. The use of smartwatches and heart rate monitors for sports, and recreational activities, as well as monitoring of health conditions, is steadily on the rise to maintain optimal physical performance\cite{pobiruchin2017accuracy}. Biological markers such as cardiac rhythm, blood pressure, and respiratory variables have been investigated using smartphones and -watches \cite{michard2021toward,liang2016wearable}, which could aid in the detection of, for instance, atrial fibrillation \cite{perez2019large}. Although useful, these technologies are limited to the estimation of bio variables by the type of sensors available for use \cite{lu2016healthcare}.

In the field of kinesiology, the estimation of moments about the joints in the human body has been vastly studied for understanding human biomechanics as well as applying them to sports science and wearable robotics. For the estimation of biological joint torques, the use of motion capture systems in conjunction with force plates has been widely utilized \cite{whittle2014gait}. This setup is highly accurate but cannot be utilized in outdoor scenarios. Therefore, wearable sensors offer several benefits over commercially available smart devices such as phones and watches. Typically, inertial measurement units have been utilized to estimate the kinematics of the human body \cite{karatsidis2016estimation}. However, several assumptions such as person-specific limb mass and the location center of mass have to be made and can be tedious to obtain due to the required complex equipment such as 3D scanners \cite{davidson2008estimating}. 

Therefore, several groups have worked on designing insole sensors for measuring ground reaction forces via the use of several types of sensors. Commercially available load cells have been utilized for integrating sensing into insoles \cite{veltink2005ambulatory} but have the downside of being bulky. A solution to overcome this downside is to use soft/flexible sensors. For instance, by using air pressure sensors embedded in the shoe \cite{kong2009gait}, capacitive sensors embedded in textiles \cite{Zhang2019}, and piezoresistive sensors\cite{de2021development}. However, these insoles were often limited to the measurement of vertical ground reaction force (GRF) and not the mediolateral and anteroposterior GRFs. Typically, these are realized using manufacturing processes such as molding \cite{kim2018air} and manually coiling a tube \cite{wachtel2017design}. More recently, researchers have been investigating 3D printing, which can realize high geometric complexity without requiring complicated processing steps.

Due to advances in additive manufacturing, it is now possible to 3D print compliant materials such as thermoplastic elastomers\cite{saari2015additive} and use conductive polymers\cite{criado2021additive} for sensing applications. Moreover, 3D printing has the potential to create patient-specific solutions, to for instance reduce foot-related musculoskeletal ailments \cite{shaikh2023effects}. Furthermore, printing with a sensorized material could allow for personalized insoles that can estimate the ground reaction forces. 

Some researchers have already investigated 3D printed insoles with integrated sensing \cite{majewski2017design,samarentsis20223d,hao20193,binelli2023digital}. One approach is to embed 3D-printed electrodes separated by a dielectric to act as a capacitive sensor \cite{samarentsis20223d,ntagios20223d,valentine2017hybrid}. Another method is to 3D print the insole and embed optical fibers \cite{hao20193} or force-sensing-resistors (FSRs) \cite{majewski2017design}. The latter approach uses piezoresistive sensing, which requires simple read-out circuitry. This piezoresistive sensing was also shown to be possible in the 3D printed soft insole \cite{binelli2023digital}. 

A challenge when using piezoresistive sensors is their inherent nonlinearities and hysteresis making the relationship between resistance change and the stress not straightforward. To convert the resistance change to a stress, researchers have investigated machine learning \cite{Choi, oubre2021estimating,yabu2023estimation} for insoles with embedded commercial force sensing resistors. This approach was shown to work for estimating the 3D ground reaction forces (GRFs) \cite{oubre2021estimating,yabu2023estimation}. Another approach is to use system identification, which can give models with (relatively) lower computational complexity. System identification of Hammerstein-Wiener models has been investigated for commercial force sensing resistors \cite{Saadeh} but also strain-sensing of 3D-printed piezoresistive sensors \cite{WillemsteinPRA}. However, 3D-printed foam-like (i.e. porous) piezoresistive sensors for 3D GRF estimation have not, to the authors' knowledge, been investigated. 

The combination of sensing and foam-like behavior makes them an interesting candidate for sensorized insoles. The foam-like structure makes such sensors programmable in terms of stiffness \cite{InFoam}, which can be used to fabricate patient-specific insoles. For example, such insoles could be used for diabetic patients to reduce foot ulcers \cite{ma2019design}, which can be augmented with foam-like sensors to provide more information. 

This work aims to investigate the feasibility of  3D printed foam-like piezoresistive for estimating 3D GRFs using identified models. These sensors should be integrated into an insole and not feel like an external element through their softness. This study can act as a stepping stone toward 3D-printed insoles that can be personalized and sensorized. Compared to other sensorized insoles in literature, the presented insoles leverage additive manufacturing of soft thermoplastic elastomers to make foam-like sensors using our InFoam Method \cite{InFoam} and employing system identification to allow for 3D GRF estimation. Thereby providing a manufacturing and calibration method for soft sensorized insoles that can estimate the ground reaction forces. Furthermore, the presented approach of combining 3D-printed foam-like structures and system identification could be useful for soft wearable force sensors.

This paper is organized as follows; Section 2 details the design of the soft sensorized insole, the force estimation method, its fabrication method, and the test procedure used to evaluate the insole. Sections 3 and 4 present the results and discussion, respectively. These sections discuss the results in terms of sensor behavior during the gait cycle and the estimation of the ground reaction forces (vertical, mediolateral, and anteroposterior) for personalized models for a single and multiple day(s). In addition, a subsection in the discussion is dedicated to comparing our insole to other insoles in literature. Lastly, the conclusions from this work and future work are detailed in Section 5.

\section{Methods}
\subsection{Soft Sensorized Insole Design \& Fabrication}

\begin{figure}[h]
\centering
\includegraphics[width=0.95\textwidth]{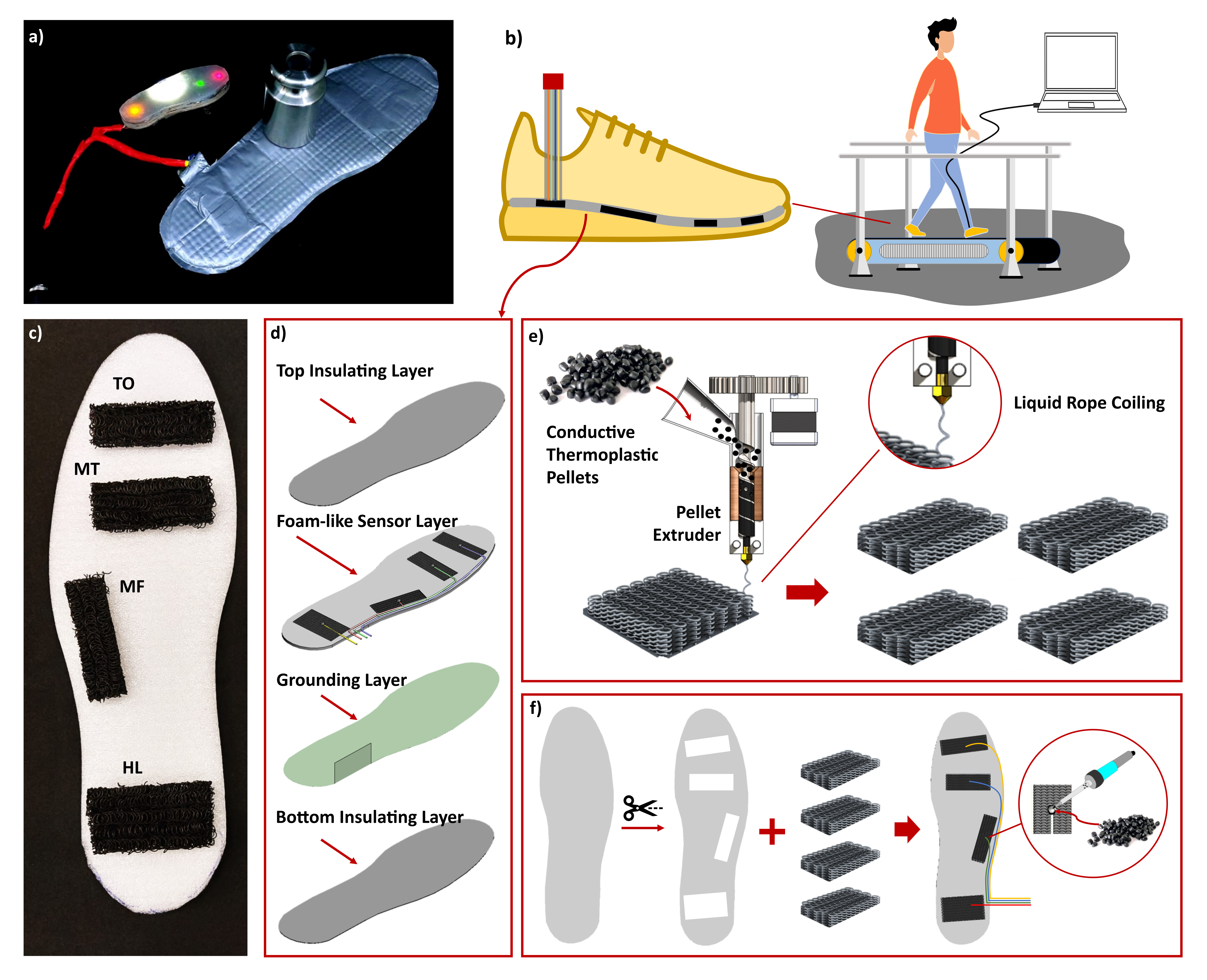}
\caption{(a) The sensorized insole while measuring the application of a force due to the foam-like sensor deforming reducing the resistance and increasing the LED's brightness, (b) we evaluated the insole of these insoles on an instrumented treadmill, (c) the foam-like sensors were located on the toe (TO), metatarsal (MT), midfoot (MF), and heel (HL). (d) the sensorized insoles are composed of four layers, (e,f) the fabrication approach for realizing the insole, which includes (e) combining 3D printing and liquid rope coiling to create foam-like structures and (f) subsequent assembly and soldering steps}
\label{fig:insoledesign}
\vspace{-10pt}
\end{figure}

Realizing a sensorized insole requires the integration of sensing into a flexible/soft foot-shaped structure. The softness of foam-like structures can be very promising for such an application. Foam-like structures have been investigated in literature and were shown to be very promising due to their large change in resistance when a force/strain is applied \cite{WillemsteinPRA}. This piezoresistive effect is also used in this work. The significant change in resistance can also be seen in Figure \ref{fig:insoledesign}(a). Due to the large change (>75\% possible, as will be shown later) in resistance, the LED can go from dim to very bright. Thereby realizing a sensitive yet soft sensor.

In this work, foam-like sensors are used for sensorized insoles to measure ground reaction forces (GRFs). Specifically, we focus on using the insoles on an instrumented treadmill (Figure \ref{fig:insoledesign}(b)). To reconstruct the GRFs, the sensors need to capture enough data during walking, To this end, the sensors were distributed over the entire length of the foot. Specifically, the sensors were added at the heel (HL), midfoot (MF), metatarsal (MT), and toes (TO) (Figure \ref{fig:insoledesign}(c)). These locations should allow for the reconstruction of aspects of the gait cycle such as heel strike and toe-off.  Furthermore, these locations experience most of the pressure on the foot during walking thereby aiding in ground reaction force estimation.

The complete insole also required the addition of insulation to separate the user and conductive components. To this end, an insole was designed consisting of four individual layers (Figure \ref{fig:insoledesign}(d)): a soft foam layer that integrates the sensor, two insulating (top and bottom) layers, and a layer with grounding electrodes. The purpose of the insulating layers was to prevent the user from creating an additional conductive path but also to package the entire structure. 

The sensing of the insole was accomplished by the grounding and sensor layer. The sensor layer consisted of four piezoresistive foam-like (porous) sensors (Figure \ref{fig:insoledesign}(c)). These piezoresistive sensors \cite{WillemsteinPRA} change their resistance when subjected to a load. The resistance of the sensors was measured using a voltage divider with the foam-like sensor inside the shoe and a second resistor external to the shoe. To complete the voltage dividers a common grounding layer was added inside the insole. A small strip was added at the edge of the common ground layer to provide a connection point for the wiring outside of the user's shoe.

These sensors need to be soft and lightweight to be acceptable for wearable sensing applications. To adhere to this requirement the sensor was made from a soft foam-like (porous) structure. The foam-like nature makes it both soft and lightweight. These porous sensors can be realized using 3D printing through our InFoam method \cite{InFoam}. This method exploits the liquid rope coiling effect, which is the coiling effect also seen when dropping viscous liquid such as honey from a certain height. The empty space of the coils creates pores leading to a structure with programmable porosity. By printing this foam-like (porous) structure out of a conductive thermoplastic elastomer, the structure can act as a piezoresistive sensor \cite{WillemsteinPRA}.

The insole was fabricated using the approach shown in Figure \ref{fig:insoledesign}(e) and (f). The first step was to print the porous sensors, which were printed using our InFoam method \cite{InFoam}.  Within this work, a modified Ender 5 Plus (Shenzhen Creality 3D Technology Co., Ltd., China) with a custom screw extruder was used. The screw extruder used TC7OEX-BLCK pellets (Kraiburg TPE, Germany) that have a Shore Hardness of 70A and volume resistivity of 10 $\Omega$cm. These were printed at 195$^{\circ}$C with a 0.6 mm nozzle. Using the conductive thermoplastic elastomer (cTPE) pellets the piezoresistive sensors were printed with sizes of 60 mm by [20, 25, 20, 30] mm porous blocks for the toe, metatarsal, midfoot, and heel sensors, respectively. The foam-like sensor had a height of 8 mm and a porosity of 73\%, which was experimentally determined to not be noticeable by the participant when integrated into the insole.

In the second step (Figure \ref{fig:insoledesign} (f)), a sheet of polyethylene foam (thickness of 5 mm) was cut to shoe size 43 (EU). Four pockets of appropriate size (see above) were cut at the toe, metatarsal, midfoot, and heel for embedding the sensors. Subsequently, the 3D-printed sensors were placed in each pocket. An electrode (a copper wire) was attached to every sensor. These wires were soldered to the sensor by using cTPE pellets. These wires were guided through the insole's edge by using tape to have a single point of entry at the lateral malleolus.

Afterwards, the common grounding layer was added. This common grounding layer consisted of a woven sheet of conductive textile (Adafruit, USA) that was connected to the Arduino. This piece of textile had a piece that stuck out near the lateral malleolus. Lastly, to finalize the insole, an insulating layer was added on top and the bottom of the insole by wrapping it in insulating tape. 

\subsection{Force Estimation by Identified Models}
\begin{figure}[h]
\centering
\includegraphics[width=0.95\textwidth]{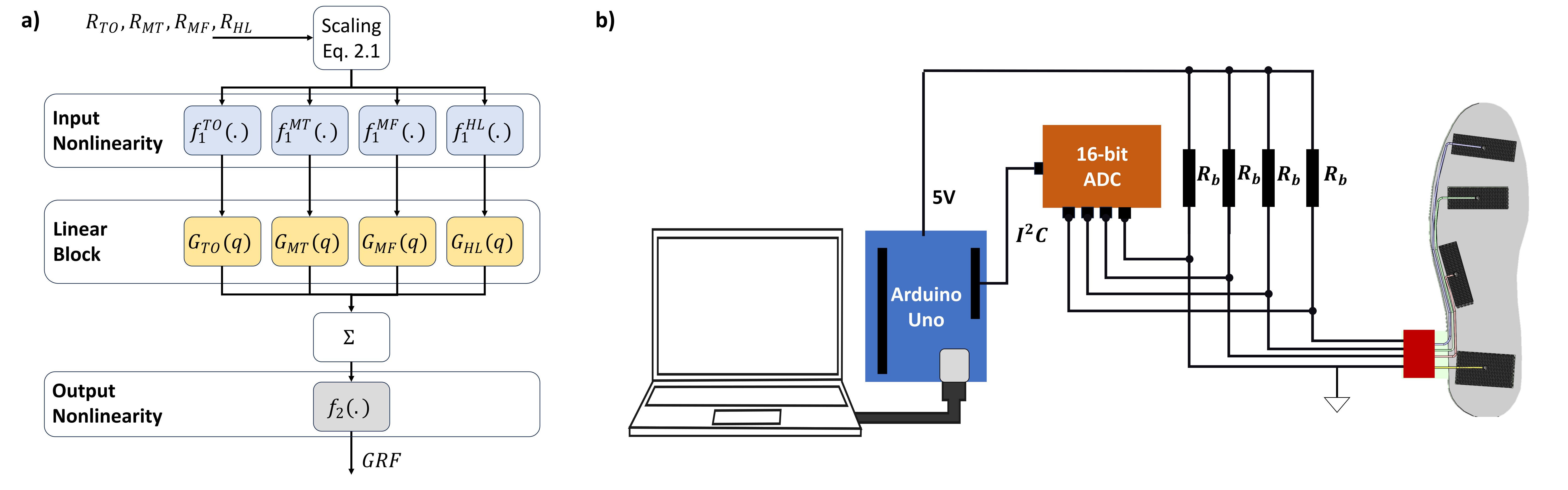}
\caption{(a) The Hammerstein-Wiener model's pipeline that converts the resistance change of the sensor to a ground reaction force (GRF), and (b) the measurement setup of the right sensorized insole using an Arduino and a four-channel 16-bit analog-to-digital converter (ADC) connected through I$^2$C, for the left insole a second ADC was added and connected in the same way to the same Arduino}
\label{fig:shoe}
 \vspace{-5pt}
\end{figure}

The ground reaction forces (GRFs) are estimated by combining the resistance change of the four piezoresistive foam-like sensors. These sensors change resistance when subjected to a force by the collapse of the porous structure reducing the resistance. Within this work, the change in resistance $\Delta R$(\%) is used for force estimation. The change in resistance at timestep $i$ is defined as (with $R_0$ the resistance when the user’s feet are in the air (i.e. zero-load))
\begin{equation}
    \Delta R(i) = \frac{R(i)-R_0}{R_0}\cdot 100\%
    \label{eq:dR}
\end{equation}
This change in resistance can be exploited to estimate the gait cycle behavior. Unfortunately, the change in resistance of piezoresistive porous sensors is nonlinear and exhibits hysteresis. These nonlinearities need to be compensated for to estimate the ground reaction forces. In literature, people solve this by using a model-based solution. Within this work, Hammerstein-Wiener (HW) models are employed as they require a relatively low number of parameters yet can still provide good force estimates, as seen in \cite{Saadeh}. The HW model is a class of nonlinear models that uses a block-oriented structure. The HW model consists of three blocks as seen in Figure \ref{fig:shoe}(a). 
 
These three blocks include two nonlinear functions ($f^j_1,f_2$) and a linear transfer function ($G_j$). This model can be expressed as:
\begin{equation}
    \hat{F}(i) = f_2\left( \sum_{j \in S} G_j(q)\left(f^j_1\left(\Delta R_j(i)\right) \right)\right)
\end{equation}

Where the variable $\hat{F}(i)$ represents the estimated ground reaction force (GRF) at timestep $i$ while $S$ is the set of sensors ($TO, MT, MF, $ and $HL$). In addition, $q$ is the shift operator such that $ R(i) q^{-n}=R(i-n)$ (i.e. $n$ samples before the current time step). Note that the nonlinear input function $f_1$ consists of four nonlinear functions (one for each sensor). The HW models are identified using MATLAB's System Identification Toolbox (The MathWorks, USA). Within this work, piecewise linear functions were used as these gave good results for strain \cite{WillemsteinPRA} and force estimation \cite{Saadeh}. The number of breakpoints was determined experimentally by trying different combinations in the range of five to ten. After initial manual model estimation, the amount was set to ten breakpoints.

Models were estimated for each GRF separately, which required the estimation of a set of three multi-input - single-output systems per insole. To determine the optimal model orders, we employed a brute-force strategy that iterated over a set of pole-zero combinations. To reduce the parameter space we set the amount of pole-zero combinations equal for all inputs. For the linear models, we explored all the causal pole-zero combinations from one to ten poles. In contrast, the HW models had a slightly limited scope in terms of pole-zero combinations based on initial manual model estimation. This manual exploration led to the restriction of one to eight poles and one to five zeros. Furthermore, only causal pole-zero combinations were considered, which were run twice to reduce the odds of getting an unfavorable initial condition.

After model estimation, the models were evaluated using their root mean squared error (RMS error) and mean absolute error (MAE) scaled by the maximum change in body-weight normalized force. These errors were defined as
\begin{equation}
    \text{RMS} = \frac{100\%}{\max(F)-\min(F)} \sqrt{\frac{\sum_{i=1}^N \left(F(i)-\hat{F}(i) \right)^2}{N}}
\end{equation}
\begin{equation}
    \text{MAE} = \frac{100\%}{\max(F)-\min(F)} \frac{\sum_{i=1}^N |F(i)-\hat{F}(i)|}{N}
\end{equation}
Within this equation, the variables $F$ and $\hat{F}$ represent the measured and estimated force, respectively. Whereas $N$ is the number of time steps. In addition, the coefficient of determination $R^2$ was computed by comparing the measured and estimated GRFs. The best model was selected based on the lowest RMS error. Furthermore, to enable comparison between multiple participants the GRFs are scaled by the body weight of the participant. 

\subsection{Sensorized Insole Testing}

Two sensorized insoles, weighing 31.1 grams each and having EU size 43, were fabricated that could be embedded in a shoe for testing. The wiring of the insole was attached to the subject by using elastic bands with velcro to ensure that the wiring did not influence the walking motion. All participants walked on a dual-belt, force-instrumented treadmill (Bertec Corporation, USA) with integrated force plates (Advanced Mechanical Technology, Inc., USA).  The instrumented treadmill was used to record the ground reaction forces of both feet. Walking trials were performed on the treadmill with both increases and decreases in speeds in a trapezoidal pattern. Specifically, the trapezoidal pattern of 2.5, 3, 3.5, 3, and 2.5 km/h was used, which each lasted approximately 50 seconds. This walking trial was performed three times.

To investigate the applicability of our sensorized insole nine participants were recruited.  Before recruitment, the experimental procedure was approved by the Natural Sciences and Engineering Sciences Ethics Committee of the University of Twente (reference: 230464). Nine participants (1 female; age: 29.2 $\pm$ 6.5, weight: 80.1 $\pm$ 10.1 kg, and shoe size 43.2 $\pm$ 1.2) with no abnormal gait pattern participated in the study after giving written informed consent.  
All participants were given the same set of insoles with one dedicated for the left and one for the right shoe. They embedded the insole in their shoes. During a single experiment, the participant performed three walking trials. The first two walking trials were used for identification. Whereas the third trial was a validation dataset for the personalized models. This experiment was performed with all participants. Four of the nine participants did this experiment over multiple days. They came back for a second and third experiment of three trials on the day after and one week after. This multiday experiment was performed to provide insight into whether the model performance stayed consistent over time. During all of the walking trials, the changes in resistance were recorded using the setup shown in Figure \ref{fig:shoe}(b) for the right insole. 

The measurement setup (Figure \ref{fig:shoe}(b)) used a voltage divider to measure the resistance of the sensor using a bias resistor $R_b$ of 560$\Omega$. The output voltage was measured using a 16-bit analog-to-digital converter ADS1115 (Texas Instruments, USA) on a breakout board (Adafruit, USA) that is connected to an Arduino Uno (Arduino AG, Italy). The Arduino streamed data over USB to a laptop for further processing. The measured voltages were then converted to resistance by
\begin{equation}
    R(i) = R_b \frac{1}{V_{in}/V_{out}(i)-1}
\end{equation}
Within this equation, the variables $V_{in}$ and $V_{out}$ represent the input (5 V from Arduino) and measured voltage, respectively. The resistance changes were then computed by using $R_b$ and Eq. \ref{eq:dR}. The changes in resistance were then used as the inputs of the to-be-identified linear and Hammerstein-Wiener models for each participant.

\section{Results}

\subsection{Sensor Behavior during Gait Cycle}

\begin{figure}[h]
\centering
\includegraphics[width=0.48\textwidth]{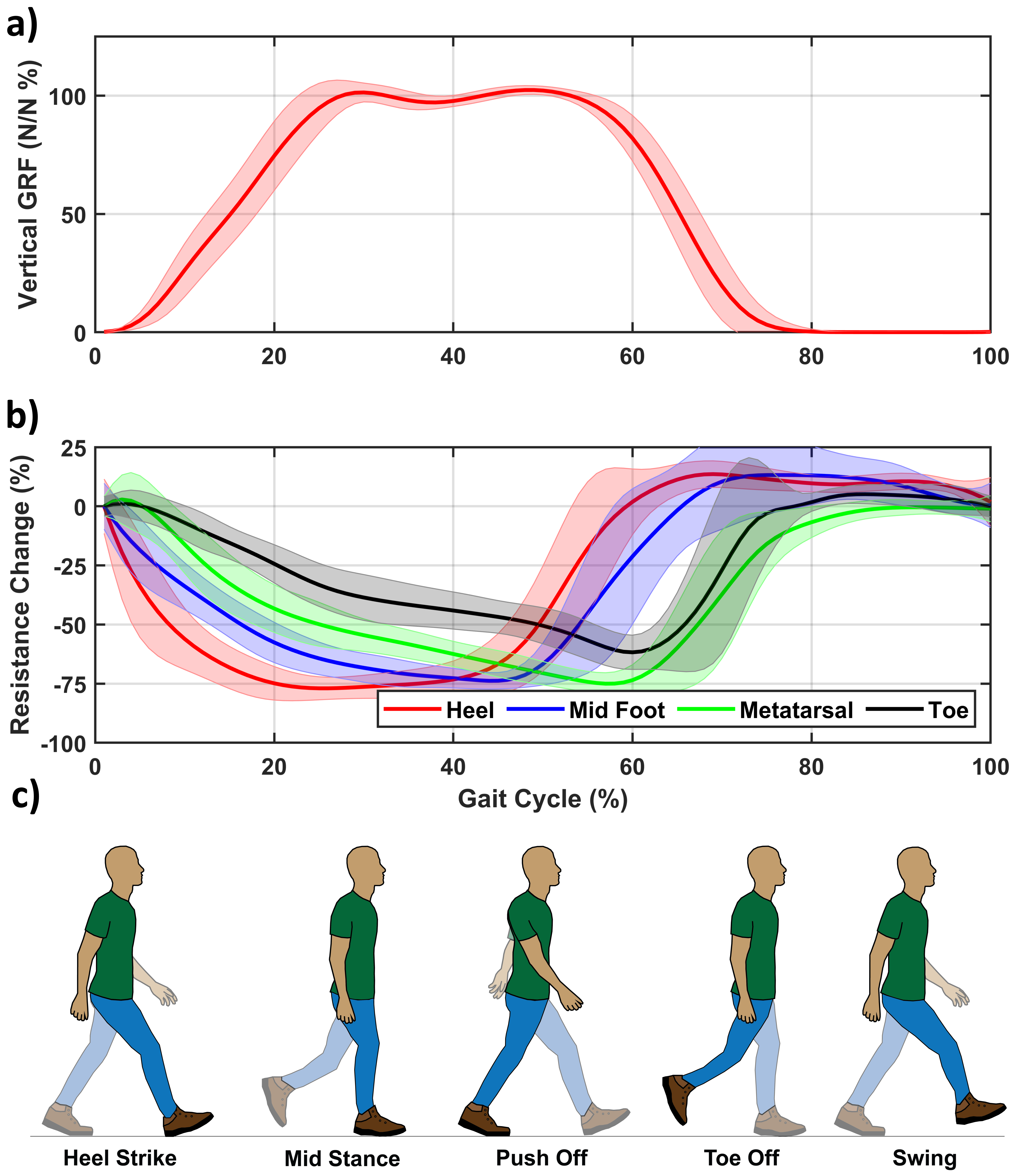}
\caption{Sensor behavior during the gait cycle of vertical forces (a), resistance change (b), and gait phase (c) for one of the participants}
\label{fig:GCE}
\end{figure}

The vertical ground reaction force (GRF) versus gait cycle is shown in Figure \ref{fig:GCE}(a,b) in error bar format based on a spline fit. A typical gait cycle can be observed with a clear heel strike ($\approx 5$\%), mid stance ($\approx 30$ to 50\%), and toe-off ($\approx 75$\%) \cite{whittle2014gait}. The resistance change curve (Figure \ref{fig:GCE}(b)) has an overall similar shape as the gait cycle but is inverted. The inverted shape is to be expected as the piezoresistive sensors decrease in resistance with increased force.

The sensor behavior over the gait cycle has similarities in their general shape to the vertical ground reaction force (GRF). The sensors reach peaks and relax in line with the change in pressure distribution expected over the gait cycle (Figure \ref{fig:GCE}(c)). From heel strike to mid-stance a successive activation from the heel to toe is seen, which indicates a gradual (expected) change in force distribution from the heel to the entire foot during mid-stance. After the mid-stance, the sensors at the midfoot, metatarsal, and toe sensor show an increase indicating that the subject is shifting weight towards his toes (as expected). Moreover, during this phase, all sensors show relaxation in the same order as the heel strike event (i.e. from heel to toe). Lastly, the shift in force towards the metatarsal/toe should be right before the toe-off, as also seen in our sensors. 

\subsection{Single Day Trials}

\begin{figure*}[h]
\centering
\includegraphics[width=0.99\textwidth]{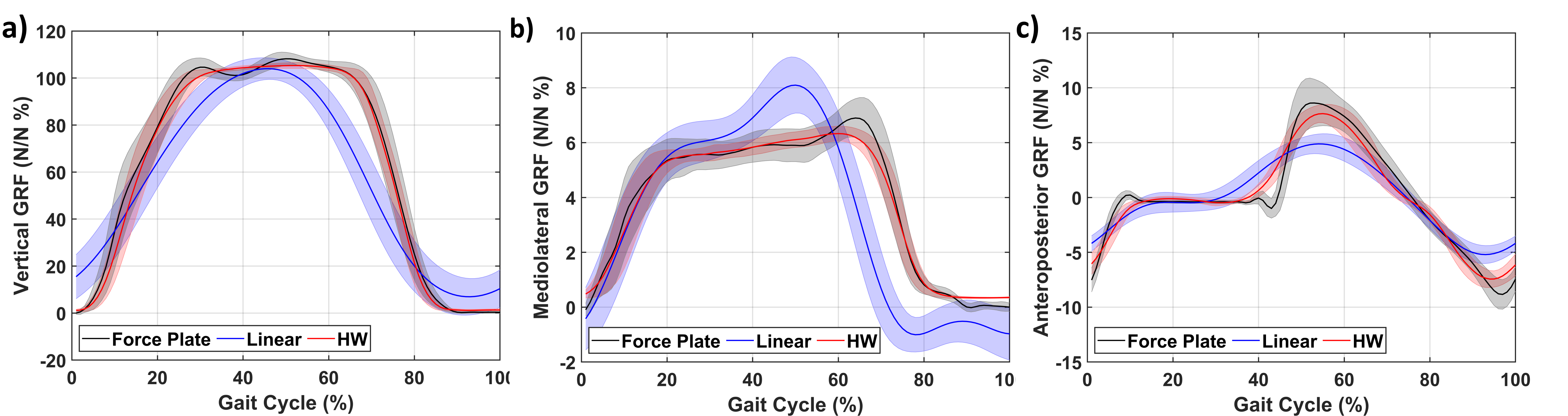}
\caption{Estimated and measured 3D ground reaction forces based on the segmented gait cycle for one of the participants for the (a) vertical, (b) mediolateral, and (c) anteroposterior forces averaged based on approximately 115 cycles on the day of model identification}
\label{fig:sdGRF}
\end{figure*}

Using the data collected during the first day of trials, a set of Hammerstein-Wiener (HW) and linear models were estimated. The estimated ground reaction forces (GRFs) by these models are shown for a segmented gait cycle (averaged over all the gait cycles during the walking trial) in Figure \ref{fig:sdGRF}(a-c). The vertical, mediolateral, and anteroposterior GRFs are decently approximated by the HW model in terms of magnitude. The linear models seem to have difficulty with the vertical GRF while the other two seem more reasonable but worse than the HW models. In general, the results for the averaged GRFs presented in Figure \ref{fig:sdGRF} show that the insole can reconstruct the GRFs' shape quite well. However, the estimated values do not show the peaks of the GRFs, which were captured by the force plate data indicating some discrepancies.  
 
The level of accuracy was further explored in terms of RMS, MAE, and the coefficient of determination ($R^2$). The estimation quality was examined in terms of both the segmented gait cycle and the entire time-series. These two perspectives were employed as segmentation is useful for clinical purposes whereas time-series estimation is more important for closed-loop control of, for instance, an exoskeleton. Segmentation provides less insight for the latter due to the averaging inherent to segmentation can obfuscate outliers. These outliers can negatively impact closed-loop control. The model selection of both was done separately to investigate the effect of the model order on either application.

\begin{figure}[h]
\centering
\includegraphics[width=0.99\textwidth]{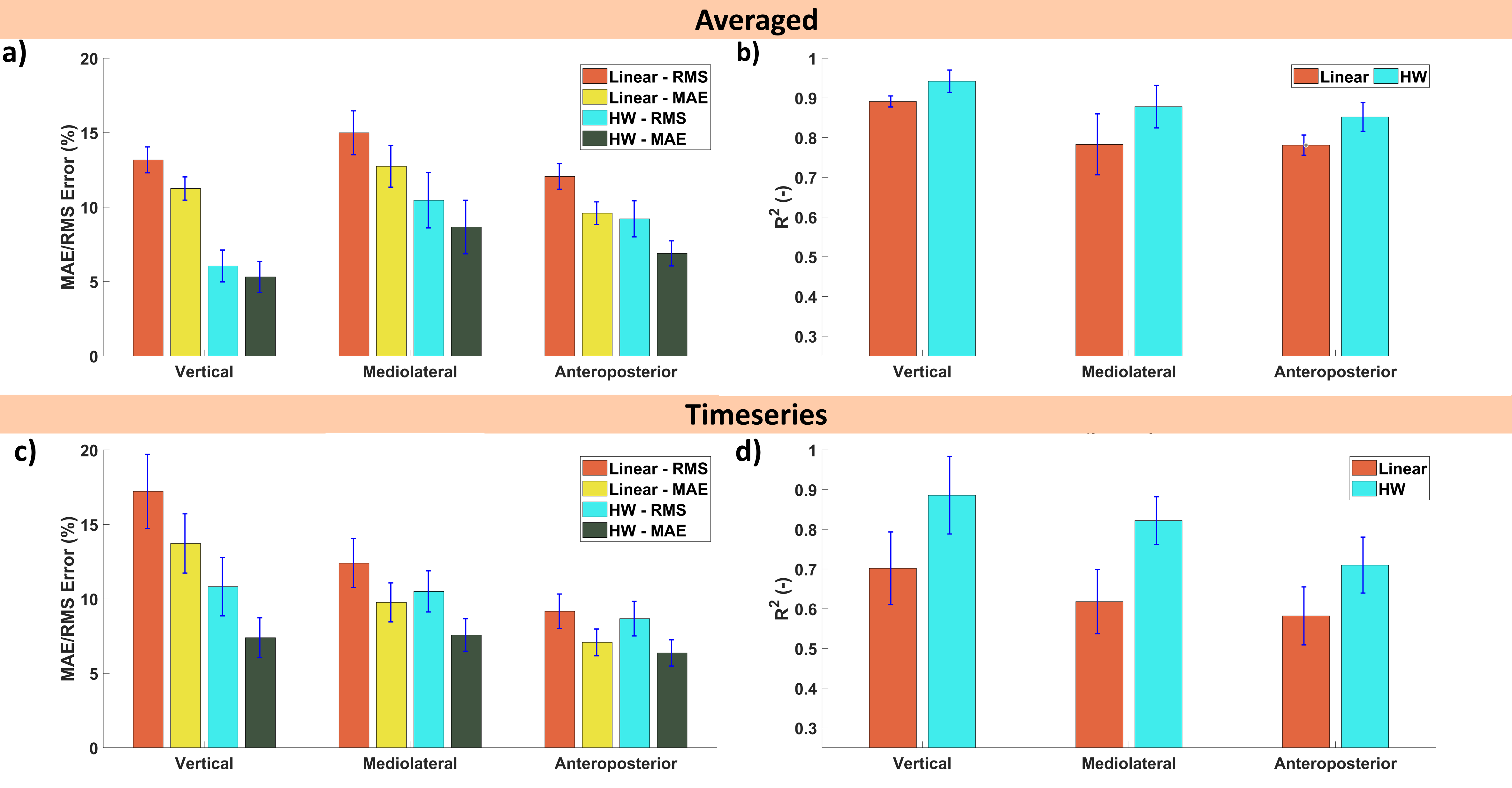}
\caption{Estimation performance on the same day as model identification with standard error of the ground reaction forces with the RMS error and MAE as a percentage of the force amplitude and the coefficient of determination $R^2$ for both the (a,b) segmented gait cycle and (c,d) time-series for the HW and linear models}
\label{fig:sdbar}
\end{figure}

Both of these results are shown in Figure \ref{fig:sdbar}. For the segmented gait cycle (Figure \ref{fig:sdbar}(a) and (b)) it can be observed that the HW model performs well with RMS errors being 6.1, 10.5, and 9.2\% for the vertical, mediolateral, and anteroposterior GRFs, respectively. Similarly, the MAE and $R^2$ also indicate good estimation performance with on average 7\% and 0.89. The results over the time-series (Figure \ref{fig:sdbar}(c,d)) show an increase in error compared to the segmented data. 
 
Looking at the results over the time-series (Fig. \ref{fig:sdbar}(c,d)), a noticeable increase in error can be observed compared to the segmented. This discrepancy is to be expected as the averaging inherent to segmentation should decrease the influence of outliers. These outliers lead to larger RMS errors whereas the MAE stays similar although the RMS error and MAE are around 10 and 7.1\%. By averaging the influence of outliers (i.e. larger mistakes in estimation) will be reduced. Such outliers would increase the RMS error more than the MAE is less affected. The $R^2$-value tells a similar story as the RMS error and still indicates reasonable explaining power especially, for the vertical and mediolateral GRFs (both exceed 0.8) while the anteroposterior has an $R^2$ of 0.7. 

These results show that the HW model outperforms the linear models by a wide margin. Due to these results, the results of the multi-day trial will focus on the HW models only. 

In addition, the same pair of insoles was re-used by the participants even though participants had different shoe sizes. However, there is no clear relationship between the shoe size and estimation performance. There are some outliers in terms of performance with smaller shoe sizes but no clear relationship can be observed, as seen by the low standard error. Implying that the usage of a single size of insole was not detrimental to estimation performance. However, the reason why it did not affect the estimation performance is not known. One reason could be that the sensors were large enough such that the sensor was still compressed even though they were not at the optimal location.

Lastly, the identified models were evaluated in terms of model order. The average number of poles and zeros are shown in Table \ref{tbl:mdlprops}. Because the amount of poles and zeros were set the same for all inputs only a single value is given. Furthermore, the amount was not rounded to preserve information. It can be observed that the order of both the linear and HW models is similar for the vertical GRF. However, for the mediolateral and anteroposterior GRF, the HW model is of higher order but with less variance. Due to our limited pole-zero combinations for model identification it is uncertain whether these pole-zero combinations are a general result. However, the poles and zeros are below the maximum indicating that the limited parameter space was sufficient in most cases. The high variance also indicates different participants required different pole-zero combinations. Furthermore, both the optimal order of the segmented and time-series have similar means and variances but no clear relationship can be observed. 

\begin{table}[htbp]
\caption{Table with the average amount of poles and zeros (with standard deviation) for the Hammerstein-Wiener (HW) and linear (L) models sorted per ground reaction force (GRF). The models were selected separately for the segmented (Seg) and time-series (TS) datasets}
\label{tbl:mdlprops}
\centering
\begin{tabular}{|c|c|c|c|c|}
\hline
    \multirow{ 2}{*}{\textbf{GRF} } & 
    \begin{tabular}{c}
    \textbf{L - Seg} \\
    \hline
    [$n_p$,$n_z$]
    \end{tabular} &
    \begin{tabular}{c}
    \textbf{L - TS} \\
    \hline
    [$n_p$,$n_z$]
    \end{tabular} &
    \begin{tabular}{c}
    \textbf{HW - Seg} \\
    \hline
    [$n_p$,$n_z$]
    \end{tabular} &
    \begin{tabular}{c}
    \textbf{HW - TS} \\
    \hline
    [$n_p$,$n_z$]
    \end{tabular} 
    \\
    \hline
    \hline
    \textbf{Vertical} & [4.2$\pm$2.4, 2.3$\pm$1.2] &[4.7$\pm$2.0, 2.8$\pm$1.6]  & [4.2$\pm$1.7, 3.7$\pm$1.6]  & [4.7$\pm$1.3, 3.3$\pm$1.2]\\ 
    \hline
    \textbf{Mediolateral} & [3.1$\pm$1.4, 2.3$\pm$1.5]  &[3.7$\pm$1.9, 2.7$\pm$1.4]   & [5.2$\pm$1.3, 2.9$\pm$1.1]& [4.9$\pm$1.8, 2.2$\pm$0.7] \\ 
    \hline
    \textbf{Anteroposterior} &[3.6$\pm$1.6, 2.9$\pm$1.8]  & [3.3$\pm$2.1, 2.3$\pm$1.5]  & [5.5$\pm$1.9, 2.7$\pm$1.3] & [6.1$\pm$2.0, 2.2$\pm$0.8] \\ 
    \hline    
\end{tabular}
\end{table}

\subsection{Multi-Day Trials}
\begin{figure*}[h]
\centering
\includegraphics[width=0.99\textwidth]{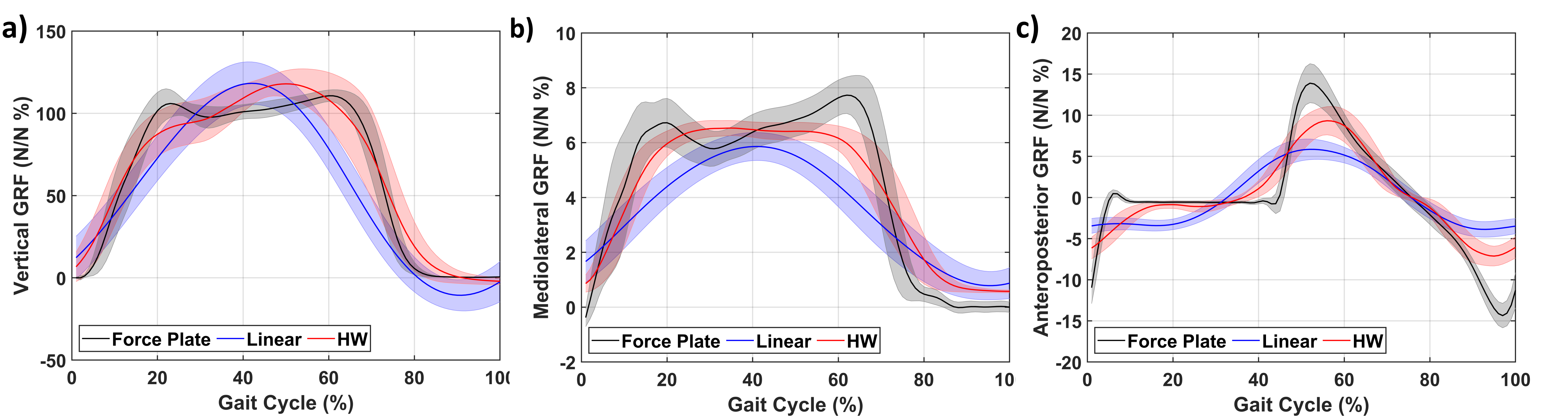}
\caption{Estimated and measured 3D ground reaction forces for the same participant as Figure \ref{fig:sdGRF} for the (a) vertical, (b) mediolateral, and (c) anteroposterior forces a week after model identification averaged based on approximately 115 gait cycles}
\label{fig:mdGRF}
\end{figure*}

The results of the single-day trial are promising. However, having to re-identify models between tests is undesirable. Therefore, the models were evaluated to see how the estimation performance behaves over time. The insoles were calibrated/identified on day 0 (i.e. from the previous subsection). Subsequently, both a day and a week after the participant wore the insoles again. The segmented gait cycle results of the GRF for a single participant are shown in Figure \ref{fig:mdGRF}. It can be observed that the models still approximate the GRF curves to a reasonable degree. However, the anteroposterior GRF does seem to have a less similar shape. Whereas the vertical and mediolateral GRF quality are less affected. Similar to the day of model identification the linear models are significantly worse than the HW model.

\begin{figure*}[h]
\centering
\includegraphics[width=0.99\textwidth]{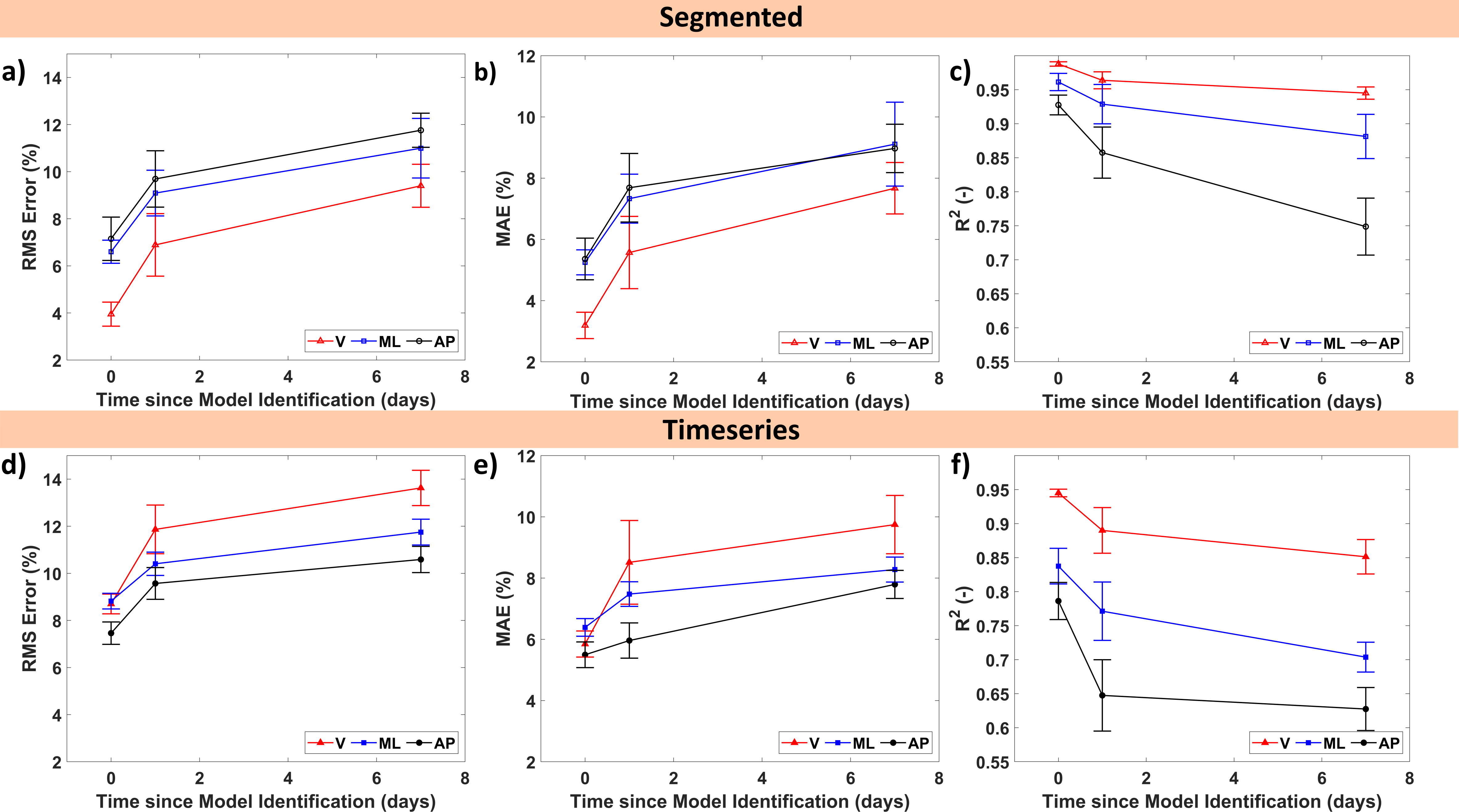}
\caption{Multi-day estimation performance with standard error of the 3D ground reaction forces with V, ML, and AP referring to the vertical, mediolateral, and anteroposterior GRF, respectively. The graphs are separated into the segmented gait cycle (a-c) and time-series (d-f) datasets. The (a,d) RMS error and  (b,e) MAE are both shown as a percentage of the force amplitude while (c,f) show the coefficient of determination $R^2$. The day of model identification is considered day 0 with the data on that day only including the four participants that are part of the multi-day trial}
\label{fig:mdbar}
\end{figure*}

The same model metrics were evaluated for both the segmented and time-series data (Figure \ref{fig:mdbar}) for the HW models. It can be observed that estimation performance decreases significantly between  day 0 and the day after for the segmented data (Figures \ref{fig:mdbar} (a-f)). In contrast, the performance on the day after and a week after seems similar for all metrics. Although the performance a week later is lower the HW models still have MAEs on average of approximately 8.6\% (of the force amplitude) and 10.7\% for the RMS error. Similarly, the $R^2$ value decreased to 0.95, 0.88, and 0.75 for the vertical, mediolateral, and anteroposterior GRFs, respectively. Thereby showing that the estimation performance has reduced but is still decent.
 
The time-series data (Figure \ref{fig:mdbar} (c,d))  gives a similar picture with the performance decreasing. A week later the HW models had a performance of 13.6, 11.8, and 10.6\% RMS error for the vertical, mediolateral, and anteroposterior GRFs, respectively. While the MAE was on average 8.6\% and the $R^2$-value also decreased to an average of 0.73 with the anteroposterior performing the poorest (0.63). Thereby indicating much poorer estimation performance but similar to the segmented gait cycle the performance drop between the day and the week after is significantly less. 

 For both the segmented and time-series data the MAE still indicates decent estimation performance (9.7\% at the most). Similarly, the $R^2$ values still show a decent value while the RMS is still below the 15\% for all GRFs. 
\section{Discussion}

This work aimed to investigate the feasibility of foam-like 3D printed sensors for estimating 3D ground reaction forces (GRFs). The overall results presented in the previous section indicate the sensors can provide a reasonable estimate of the GRFs when fed into identified Hammerstein-Wiener (HW) models. This section will discuss the results in the context of GRF estimation and other insoles in the literature. 

\subsection{Ground Reaction Force Estimation}

Both the sensor behavior and GRF estimation performance (Figures \ref{fig:GCE} to \ref{fig:mdbar}) show promise. The raw sensor data shows a correlation between the sensor behavior and the gait cycle. This result implies that the raw sensor data can be used to estimate the gait cycle. Furthermore, the correlation with the gait cycle could indicate that the GRFs (also related to the gait cycle) can be estimated.

Similarly, the averaged GRFs (Figures \ref{fig:sdGRF} and \ref{fig:mdGRF}) correlate well with the vertical GRF. This correlation can be exploited to estimate the gait phase using the estimated models. Besides estimating the gait phase from the vertical GRFs, the estimation performance of the GRFs seems reasonable (RMS error on average below 10\%). The separation in time-series and segmented does seem warranted as the averaged data performs much better. Indicating that the averaging indeed seems to obfuscate outliers, which can be detrimental to closed-loop control. This result implies that when evaluating the estimation quality of the insoles, one does need to take into account whether the time-series or segmented suffices for the intended application. Furthermore, the results imply that the HW model is significantly better at providing force estimates than linear models. 

In general, the degradation over time is notable. However, the results imply that the estimation performance degrades the most from 0 to 1 day after model identification. Evaluation over more days would be required to investigate whether the HW model's estimation performance stabilizes or further degrades. Furthermore, the segmented gait cycle and time-series both show similar qualitative degradation. However, the segmented gait cycle does seem to degrade more significantly as ratiometrically its increase seems larger.

Solving this degradation over time would make the proposed sensors more feasible. The reason why the performance degrades is not fully understood. A possible reason is that the piezoresistive sensors drift. This drift is also seen in commercial force sensing resistors, which is hypothesized to be due to the carbon particles and polymer matrix acting as a resistor-capacitor circuit \cite{paredes2018self}. This capacitive drift can charge the sensor over the course of a day changing the sensor behavior. For instance, by changing the value when no-load is applied \cite{WillemsteinPRA}. Such behavior could explain why the change between the day and the week after is less significant as the majority of walking trials were in the first week. Leading to a more significant charge on days 0 and 1 since model identification but less a week later. Such drift could be remedied by, for instance, a drift compensation circuit such as proposed in \cite{paredes2018self}.

Two limitations in this study are the use of personalized models and the evaluation of estimation performance over a week for a walking task. The former means that models needed to be identified nine times. If the models could be generalized the entire identification process would be simplified. However, for patient-specific insoles, the personalized approach could be more logical. In addition, the focus on only a walking task limits the applicability of results. In a future work, more complex tasks such as walking up stairs and jumping could be added to see how the HW models generalize.

Lastly, although the HW models provide good estimates the structure is likely not optimal. Specifically, the choice to keep the same amount of poles and zeros for each input could be sub-optimal. Further investigation could be performed by identifying a state space implementation of the HW model, which could lead to models with a more optimal combination of poles and zeros. This change could lead to lower model complexity and better GRF estimates. 

\subsection{Insole Comparison}

The developed insoles are capable of estimating the 3D ground reaction forces (GRFs) using the personalized identified models. To put the acquired model metrics into context the insole will be compared to several other force-sensing insoles. The focus will primarily be on force sensing resistors (FSRs) as these are also based on piezoresistive sensing.

Firstly, in terms of performance, the presented insoles can estimate the gait phase by using the estimated GRFs. Thereby making our insoles on par with insoles such as \cite{ Zhang2019,binelli2023digital,beccatelli} that can provide gait cycle data. However, our insoles can also estimate GRFs. A table comparing our results with four other insoles is shown in Table \ref{tbl:comparison}. Only the scaled RMS is used, as this makes comparison possible regardless of subject. Furthermore, the data of the segmented gait cycle is used as this is often used in other works as well whereas none of the works compare over multiple days so that comparison was not possible.

\begin{table*}[h]
\caption{Table with selected works of sensorized insoles. Both the RMS and $R^2$ display (in order)  the vertical, mediolateral, and anteroposterior GRF results. Treadmill refers to an instrumented treadmill with embedded force plates}
\label{tbl:comparison}
\begin{center}
\begin{tabular}{|p{2cm}||p{2cm}| p{2cm} | p{2cm}| p{2cm}| p{2cm} |}
\hline
\textbf{Metric} & \textbf{Our Insole}  & \textbf{FSR}\cite{Choi} & \textbf{FSR and angle sensor} \cite{oubre2021estimating}  & \textbf{Pneumatic} \cite{wachtel2017design} & \textbf{FSR and piezoelectric} \cite{yabu2023estimation}\\
\hline
Sensor & 3D printed & Commercial & Commercial & Custom   & Commercial\\
\hline
Number of sensors & 4 & 6 & 6 & 4 & 8\\
\hline
Principle & Piezoresistive & Piezoresistive & Piezoresistive and knee angle sensor& Pressure   & Piezo-resistive and electric\\
\hline
Model & HW & Artificial Neural Network & Random Forest & Hysteresis Compensator & Support Vector Regression\\
\hline
$R^2$ (-) & 0.94, 0.88, 0.85 & 0.96, n/a, n/a & 0.91, 0.59, 0.91 & 0.98*, n/a, n/a & not reported\\
\hline
RMS Error (\%) & 6, 10, 9 & 5, n/a, n/a & 7, 14, 7 & 32, n/a, n/a & 8, 9, 9\\
\hline
Data Acquisition & Treadmill & Treadmill + Force plate & in-ground Force plate & Treadmill & Treadmill\\
\hline
\end{tabular}
\end{center}
\end{table*}

Comparatively, our sensors are printable whereas the other insoles relied on the assembly of commercial FSRs. This structural integration should make it less noticeable to the user. Furthermore, the ability to 3D print the sensor allows for customization that can be beneficial for patient-specific applications such as those with musculoskeletal ailments \cite{shaikh2023effects} or diabetic patients \cite{ma2019design}. The ability to print a foam-like structure that can be mechanically programmed \cite{InFoam} to provide mechanical support where needed while also acting as a sensor can be useful for patient-specific insoles with integrated sensing. Furthermore, the printing of an insole would require around 32 grams of material. Assuming a cost of 5-100 euro/kg for elastomeric pellets, the total price would range from 0.37 to 6.20 euros for both insoles. This cost although preliminary indicates that material-wise it can be an affordable yet patient-specific solution.

Moreover, being lightweight and comfortable is important for user acceptance. Due to our insoles being composed of a soft foam-like sensor comfort can be achieved. Lightweight also seems attainable as our insole weighed $\approx$31.1 grams (excluding electronics). This value is difficult to put into context, as other researchers often do not report the weight of the insole. An exception is \cite{yabu2023estimation} but they only report the weight of the total package, not the insoles.

All insoles used different models for estimation. This difference makes absolute comparisons of the metrics difficult as the comparison is between the combination of a modeling/learning/identification approach and sensor not merely the sensor. 

It can be observed that our insoles perform comparably in terms of the coefficient of determination $R^2$ for the vertical GRF. The insole of \cite{wachtel2017design} comes near our $R^2$ but was only based on their repeatability testing, not GRFs. Furthermore, their evaluation was based on the time series data, which is still worse than ours. The scaled RMS error values are in the same order of magnitude for all the piezoresistive sensors.

The mediolateral and anteroposterior GRFs were also estimated in \cite{oubre2021estimating}. Their $R^2$ is lower than ours just like their RMS error for the mediolateral GRF. Whereas their anteroposterior force estimation \cite{oubre2021estimating} is better compared to ours. In contrast, \cite{yabu2023estimation} is similar to ours for all RMS errors but did not report $R^2$.

Overall, our printed sensors estimate the GRFs comparable to other insoles. Even though our insole uses two sensors less than the other two FSR-based insoles. Due to the comparable performance, our insole combined with system identification seems a promising candidate for further investigation. One aspect that is comparatively missing is estimating the center of pressure (done in \cite{Choi,oubre2021estimating}) and pressure distribution. Both can be a topic for future research.

In addition, the data acquisition method is an important point of discussion. All the insoles in Table \ref{tbl:comparison} use a set of flexible or soft sensors that compress and bend. Their raw data is then fed into a model that converts it to an estimate of the GRFs. The advantage of this approach is that the sensor design is relatively straightforward. Unfortunately, it also means that a ground truth is required to train a model.

This ground truth requires complex systems such as force plates. This requirement makes it more difficult to apply these insoles in practice. One possible solution is to use multi-axial sensors such that the models can be acquired externally, which increases sensor complexity but could simplify calibration. However, the distributed nature of the forces over the foot could make this non-trivial. For instance, in \cite{yabu2023estimation} they used both FSRs and piezoelectric film sensors combined with support vector regression for GRF estimation and saw an improvement. Another approach used by \cite{oubre2021estimating} is to complement the model with user data such as shoe size and body weight for model generalization. However, it was not reported how the model generalizes to new insoles or the training would need to be performed again. Such approaches could also be investigated in the context of 3D printed (foam-like) sensors to benefit from personalization and improved estimation as part of future work. Although our results do imply that the personalized models can outperform the generalized models.

\section{Conclusion}

Sensorized insoles are a useful tool to gain insight into aspects related to gait and health monitoring. For such insoles to be used in daily life they need to be, among others, comfortable and lightweight. Within this work, the feasibility of 3D-printed foam-like sensors for sensorizing insoles was investigated. The soft and foam-like nature of these sensors makes them comfortable and lightweight. 
\vspace{5pt}

Our findings indicate that the 3D-printed foam-like sensors combined with identified personal Hammerstein-Wiener models can estimate all three ground reaction forces with reasonable accuracy (RMS$\leq$10\% on the day itself) and decent accuracy a week later (on average 11\% and $\leq$13\%). With performance comparable to FSRs combined with machine learning. This similarity in estimation performance supports the feasibility of this approach. Furthermore, the estimation consistency over multiple subjects indicates that the combination of 3D printing, and system identification is an interesting opportunity for wearable sensors. 

\vspace{5pt}

A limitation of the presented approach is the reliance on an instrumented treadmill. Therefore, these insoles could benefit from the development of a simplified calibration method to make usage much less complex. Furthermore, future research could investigate the behavior of these insoles over longer periods and compensation strategies to preserve estimation performance. Additionally, 3D printing of the complete insole based on the 3D scan of the user's feet can be explored to evaluate the effect of customization on estimation performance and/or for creating patient-specific sensorized insoles. The current insole could also benefit from sensors specialized in measuring a specific GRF. These sensors could improve estimation performance by being only sensitive to a single force instead of all simultaneously. Such sensors could also aid in estimating the centers of pressure and visualizing the force distribution. Lastly, other movements such as running or walking up stairs could be investigated to evaluate the estimation performance in more scenarios.

\paragraph{Funding Statement}
This work was partially funded by the 4TU Dutch Soft Robotics Program.

\paragraph{Acknowledgements}
The authors would like to thank Kraiburg TPE GmbH \& Co. KG for providing the materials used in this research free of charge.

\paragraph{Competing Interests}
The authors declare none.

\paragraph{Data Availability Statement}
The datasets generated during and/or analyzed during the current study are available from the corresponding author upon reasonable request. 

\paragraph{Ethical Standards}
The authors assert that all procedures contributing to this work comply with the ethical standards of the relevant national and institutional committees on human experimentation and with the Helsinki Declaration of 1975, as revised in 2008.

\paragraph{Author Contributions}
Conceptualization: N.W., A.S.; Methodology: N.W., S.S., H.K.; Fabrication: N.W., A.S.; Data curation and Processing: N.W. Data visualization: N.W., S.S.; Data Gathering: N.W., S.S.; Writing: all authors. All authors approved the final submitted draft.

\bibliographystyle{IEEEtran}
 \bibliography{references.bib}

\end{document}